\documentclass[letterpaper, 10 pt, conference]{ieeeconf}  % Comment this line out if you need a4paper

\IEEEoverridecommandlockouts                              % This command is only needed if 
\usepackage{graphics} % for pdf, bitmapped graphics files
\usepackage{graphicx}
\usepackage{float}
\usepackage{booktabs} % for creating tabular
\usepackage{array} % for creating tabular
\usepackage{caption} % for captioning tabular
\usepackage{adjustbox}
\usepackage[utf8]{inputenc} % accents on letters
\usepackage{amssymb}  % assumes amsmath package installe
\usepackage{balance}
\usepackage{hyperref}

\usepackage{geometry}
\newcommand{\head}[1]{\textnormal{\textbf{#1}}}

\usepackage{color}
\definecolor{markupcolour}{rgb}{0.7,0.0,0.1}

\graphicspath{{./img/}}

\title{\LARGE \bf
Recognizing Objects In-the-wild: Where Do We Stand?
}

\author{Mohammad Reza Loghmani$^{1}$, Barbara Caputo$^{2}$ and Markus Vincze$^{1}$% <-this % stops a space
\thanks{$^{1}$Mohammad Reza Loghmani and Markus Vincze are with the Vision4Robotics Group, Automation and Control Institute (ACIN), TU Wien, Vienna, Austria
        {\tt\small [loghmani, vincze]@acin.tuwien.ac.at}}%
\thanks{$^{2}$Barbara Caputo is with the VANDAL Laboratory, Italian Institute of Technology, Milan, Italy
        {\tt\small Barbara.Caputo@iit.it}}%
}

\begin{document}

\maketitle
\thispagestyle{empty}
\pagestyle{empty}

%%%%%%%%%%%%%%%%%%%%%%%%%%%%%%%%%%%%%%%%%%%%%%%%%%%%%%%%%%%%%%%%%%%%%%%%%%%%%%%%
\begin{abstract}

The ability to recognize objects is an essential skill for a robotic system acting in  human-populated environments. Despite decades of effort  from the robotic and vision research communities, robots are still missing good visual perceptual systems, preventing the use of autonomous agents for real-world applications. The progress is slowed down by the lack of a testbed able to accurately represent the world perceived by the robot in-the-wild. In order to fill this gap, we introduce a large-scale, multi-view object dataset collected with an RGB-D camera mounted on a mobile robot. The dataset embeds the challenges faced by a robot in a real-life application and provides a useful tool for validating object recognition algorithms. Besides describing the characteristics of the dataset, the paper evaluates the performance of a collection of well-established deep convolutional networks on the new dataset and analyzes the transferability of deep representations from Web images to robotic data. Despite the promising results obtained with such representations, the experiments demonstrate that object classification with real-life robotic data is far from being solved. Finally, we provide a comparative study to analyze and highlight the open challenges in robot vision, explaining the discrepancies in the performance.

\end{abstract}

%%%%%%%%%%%%%%%%%%%%%%%%%%%%%%%%%%%%%%%%%%%%%%%%%%%%%%%%%%%%%%%%%%%%%%%%%%%%%%%%
\section{INTRODUCTION}

Objects are ubiquitous in our everyday lives. Every common activity, such as cooking or cleaning, implies the capability of understanding and operating a set of objects to successfully complete a task. In order for a Service Robot (SR) to operate in human environments as well, the ability to recognize objects is a basic requirement. Object recognition is rarely a self-contained task, but it is rather a proxy for a large variety of high-level tasks, such as navigation, manipulation and user interaction, that heavily rely on an accurate description of the visual scene.

The advent of deep learning has had a huge impact on the object recognition task after decades of plateaued results. The progressive design of deeper and more sophisticated networks, starting from AlexNet~\cite{alexnet}, VGG~\cite{vgg}, Inception~\cite{inception}~\cite{inception-v2-3}~\cite{inception-v4} to ResNet~\cite{resnet}~\cite{wide_resnet}, has led to outstanding results in competitions such as the ImageNet Large Scale Visual Recognition Challenge (ILSVRC)~\cite{ilsvrc}. Arguably the primary driving force of the deep learning revolution is the availability of large scale datasets. The majority of these datasets, such as the popular ImageNet~\cite{imagenet}, Pascal VOC~\cite{pascal_voc}, and Caltech-256~\cite{caltech-256}, are composed of images collected through Web search engines. However, the representation of the visual world provided by these datasets implies a bias from the observer (a human photographer) and the Web search engines \cite{unbiased} that are incoherent with the representation perceived by, for example, a SR. It is then legitimate to ask whether the features learned from Web-based datasets can generalize well to robotic data, despite the aforementioned bias.

   \begin{figure}[t]
      \centering
      \includegraphics[width=0.45\textwidth]{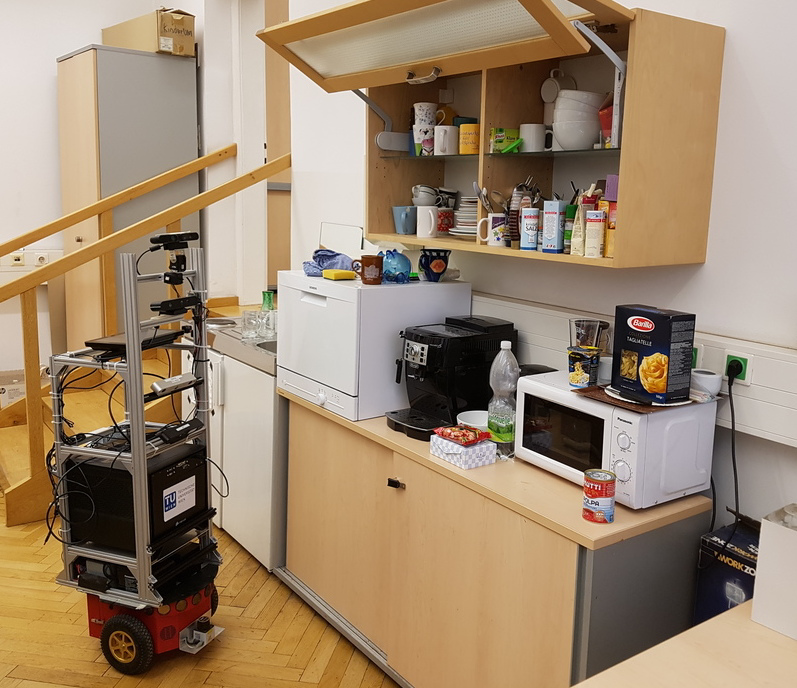}
      \caption{Glimpse of the data collection process with the robotic platform (left) acquiring data of a cluttered scene populated with everyday objects.}
      \label{fig: v4core}
   \end{figure}

%Repeatability and reproducibility are the pillars of scientific experiments. The availability of standard references to benchmark new methods and replicate published results has determined the consolidation and growth of different scientific fields.

In the last years, computer vision has progressed enormously due to the establishment of standard references and benchmarks, e.g. ImageNet, which have enabled consistent comparison and development of new methods. Unfortunately, the robot vision community has not experienced the same progress due to the lack of accurate testbeds for validating novel algorithms. In the past years, the RGB-D Object Dataset~(ROD)~\cite{wrgb-d} has become ``de facto" standard in the robotic community for the object classification task \cite{obj_rec} \cite{conv_rec} \cite{depthnet}. Despite its well-deserved fame, this dataset has been acquired in a very constrained setting and does not present all the challenges that a robot faces in a real-life deployment. In order to fill the existing gap in the robot vision community between research benchmark and real-life application, we introduce a large-scale, multi-view object dataset collected with an RGB-D camera mounted on a mobile robot (see figure~\ref{fig: v4core}), called Autonomous Robot Indoor Dataset~(ARID). The data are autonomously acquired by a robot patrolling in a defined human environment. The dataset presents 6,000+ RGB-D scene images and 120,000+ 2D bounding boxes for 153 common everyday objects appearing in the scenes. Analogously to ROD, the object instances are organized into 51 categories, each containing three different object instances. In contrast, our dataset is designed to include real-world characteristics such as variation in lighting conditions, object scale and background as well as occlusion and clutter. To our knowledge, no other robotic dataset embedding all the challenges of real-life data can be found in the literature. All the collected data, together with the information needed to replicate the experiments, is publicly available at \url{https://www.acin.tuwien.ac.at/en/vision-for-robotics/software-tools/autonomous-robot-indoor-dataset/}.

In addition to introducing a new dataset, we inspect the effectiveness of features learned from the Web domain on robotic data and compare them with the features learned from the RGB-D domain. This comparison is made possible by collecting a second dataset containing the images downloaded from the Web representing the same categories as ARID. The acquisition of this Web-based dataset is performed by using query expansion strategies from \cite{web_download} on different search engines followed by a manual cleaning to remove noisy images. Exhaustive experiments with different deep convolutional networks demonstrate that, despite the greater similarity between the RGB-D and the robotic domain, models learned from Web images are more effective. Finally, the best performing network, ResNet-50, is used to study the classification results on subsets of ARID representing three problematic characteristics of robotic data: small images, occlusion and clutter. The experiments point out small images as the main challenge of robotic data, indicating a path to follow for the resolution of the object classification problem for robotics.

In summary, our contributions are the following:
\begin{itemize}
\item a new RGB-D object dataset, collected in-the-wild with a mobile robot, that  provides a ``litmus test" for the validation of object recognition algorithms developed for robotic applications,
\item a detailed analysis of publicly available RGB-D datasets from a robotic perspective, 
\item comprehensive experiments with several well-established deep convolutional networks, comparing the effectiveness of data coming from the Web and RGB-D domain in generating features for object classification in robotics, and
\item a study of the main factors responsible for the difficulties faced by classifiers on robotic data.
\end{itemize}

The rest of the paper is organized as follows: the next section positions our approach compared to related work, section~\ref{sec: arid} introduces the proposed robotic dataset,  section~\ref{sec: experiments} presents the experimental results and section~\ref{sec: conclusions} draws the conclusions.

\begin{table*}[t]
\captionof{table}{Summary of the characteristics of different RGB-D datasets with focus on variation in lighting condition, variation in scale, multiple views, occlusion, clutter, variation in background and whether or not the data are collected directly from a robot. \textit{Not Available} (NA) indicates that the dataset focuses on object instances rather than categories and the number of categories is unknown.}
\label{tab: datasets}
%\fontsize{11}{9}\selectfont
\centering
\bgroup
\def\arraystretch{1.5}%  1 is the default, change whatever you need
\begin{tabular}{@{}*9l@{}}
 \toprule[1.5pt]
\multicolumn{2}{c}{\head{Dataset}}
& \multicolumn{6}{c}{\head{Characteristic}}\\
 \head{Name} & \head{\# classes} &
 \head{light var.} & \head{scale var.} & \head{multiview} & \head{occlusion} & \head{clutter} & \head{bkg var.} & \head{robot}\\
 \cmidrule(l){1-2}\cmidrule(l){3-9}
 RGB-D Object Dataset~\cite{wrgb-d} & 51 & %
&  & \checkmark &  &  &  & \\
 RGB-D Scene Dataset~\cite{wrgb-d_scene} & 5 & %
  & \checkmark & \checkmark & \checkmark & \checkmark & \checkmark &  \\
  BigBIRD~\cite{bigbird} & NA & %
 &  & \checkmark &  &  &  & \\
  Active Vision Dataset~\cite{activevision} & NA & % 
 \checkmark & \checkmark & \checkmark & \checkmark &  & \checkmark & \checkmark \\
  JHUIT-50~\cite{jhuit-50} & NA & %
&  & \checkmark &  &  &  & \\
  JHUScene-50~\cite{jhuscene-50} & NA & %
&  & \checkmark & \checkmark & \checkmark &  & \\
  iCubWorld Transf.~\cite{icubwt} &  15 & %
 \checkmark & \checkmark & \checkmark &  &  & \checkmark & \checkmark \\
 \textbf{Autonomous Robot Indoor Dataset (ARID)} &  51 & 
 \checkmark & \checkmark & \checkmark & \checkmark & \checkmark & \checkmark & \checkmark \\
 \bottomrule[1.5pt]
\end{tabular}
\egroup
\end{table*}

\section{RELATED WORK}
\label{sec: related_work}

In the following, we first analyze the characteristics of existing RGB-D datasets from a robotic perspective. Then, we review related works on the transferability of learned features across different domains by focusing on the Web and RGB-D domains. 

\subsection{Datasets}
\label{subsec: rw_datasets}

During the last decade, a variety of datasets have been made publicly available for research. With the popularization of deep neural networks, which require a considerable amount of data for training, the race for large-scale datasets has become more intense. While Web images exist in abundance, robotic images are difficult to obtain because platforms are expensive and data acquisition is time consuming. Nevertheless, the robotic community has produced some interesting datasets. In particular, for indoor objects, the most relevant datasets are JHUIT-50, BigBIRD, iCubWorld Transformation, ROD, and the Active Vision Dataset.

ROD~\cite{wrgb-d} contains 300 objects from 51 categories spanning from fruit and vegetables to tools and containers. Despite the availability of multiple views, each object is presented in isolation and variation in lighting condition, object scale and background are missing. The corresponding scene dataset, the RGB-D Scene Dataset~\cite{wrgb-d_scene}, presents multiple objects in the same scene, but considers only five object categories.
BigBIRD~\cite{bigbird} contains 125 common human-made objects, with particular focus on boxes and bottles. This dataset is specifically designed for instance recognition and the selected objects belong to very few categories. In addition, occlusion, clutter, scale and light variation are not captured. A more recent dataset, the Active Vision Dataset~\cite{activevision}, uses a subset of 33 objects from BigBIRD in densly acquired scenes. The data is directly acquired by a robot and it embeds most of the nuisances typical of real-life data. Nevertheless, the limited number of considered objects makes this dataset unsuitable for classification.
JHUIT-50~\cite{jhuit-50} contains 50 industrial objects and hand tools used in mechanical operations. The objects are captured in isolation and from multiple viewpoints. Due to its limited scope, this dataset is more suitable for instance recognition rather that classification. In addition, nuisances such as occlusion, clutter, scale and light variation are not captured. The corresponding scene dataset, JHUScene-50~\cite{jhuscene-50}, includes occlusion and clutter, but limits the number of considered objects to 10. 
iCubWorld Transformation~\cite{icubwt} contains 150 common indoor objects from 15 different categories. The data are collected directly with the iCub humanoid robot~\cite{icub}. This dataset addresses specifically variance in the background as well as the variance in scale and rotation of the object. Nevertheless, each object is presented in isolation, avoiding problems caused by cluttered scenes. 

Despite the high-quality that characterizes each of these datasets, their constrained setting makes them incoherent with real-life data. In addition, only the Active Vision Dataset and the iCubWorld Transformation present data collected directly from a robot. Table~\ref{tab: datasets} presents a summary of the characteristics of the datasets discussed above and highlights that, differently from other datasets, ARID embeds all these characteristics.

\subsection{Transfer Learning}
Deep convolutional networks are currently dominating several computer vision tasks. One of the key factors contributing to their success is the transferability of the produced deep representation for a variety of visual recognition tasks. The deep representations, also called features, of these networks have been empirically proven to be superior to traditional hand-crafted features, e.g.~\cite{tl1}~\cite{tl2}~\cite{tl3}. In order to take advantage of the generalization power of deep models, the networks need a large amount of training data. For this reason, large-scale datasets, such as ImageNet, with millions of samples, have been extensively used across different domains. It is common practice to further adapt the deep representation learned from a large dataset to the specific domain of interest through fine-tuning \cite{ft1} \cite{ft2}, i.e.,~by refining the representation using annotated data from the novel task in a subsequent training stage.

The effectiveness of using features learned from the Web domain in the robotic domain has been previously studied~\cite{web_download}~\cite{teaching_icub}. In Massouh et al.~\cite{web_download}, features learned from the Web domain are tested on the RGB-D Object Dataset, while in Pasquale et al.~\cite{teaching_icub}, features learned from ImageNet are used to train a classifier on the iCubWorld28~\cite{teaching_icub}, a former version of the iCubWorld Transformation. Although both works exhibit interesting results, we claim that, due to the intrinsic constraints discussed in section \ref{subsec: rw_datasets}, the utilized datasets cannot be considered as reliable representatives of real-life robotic data. In addition, only AlexNet and Inception are used to produce the analyzed features. Our work exhaustively benchmarks deep models obtained with five different networks against a robotic dataset collected in-the-wild.

%Recently, fine-tuning pre-trained networks towards new target datasets has become a common practice, since it is more effective and allows a faster convergence than training a deep network from scratch. When fine-tuning, the first \textit{n} layers from a pre-trained network are directly copied to the target network, while the rest of the layers are randomly initialized. At this point, all or a subset of the layers are further adapted, or fine-tuned, to the target dataset, usually selecting a smaller learning rate. Yosinski et al.~\cite{ft2} and Chu et al.~\cite{ft1} provided extensive studies on different datasets to identify the best practices for fine-tuning. Both these researchers agree on copying and fine-tuning all the pre-trained layers to the target network (except for the classification layer, since it is task-dependent) as the best practice, given a sufficient amount of target data. For our experiments, we rely on these guidelines and fine-tune all the layers, except for the classification one, with an appropriate learning rate.

\section{AUTONOMOUS ROBOT INDOOR DATASET}
\label{sec: arid}

In the following, we describe the characteristics of the proposed robotic dataset by highlighting its most significant peculiarities. In addition, we unveil the protocol used for the autonomous data collection and the details of the provided annotation.

\subsection{Scope and Motivation}
\begin{figure*}[t]
      \centering
      \includegraphics[width=0.8\textwidth]{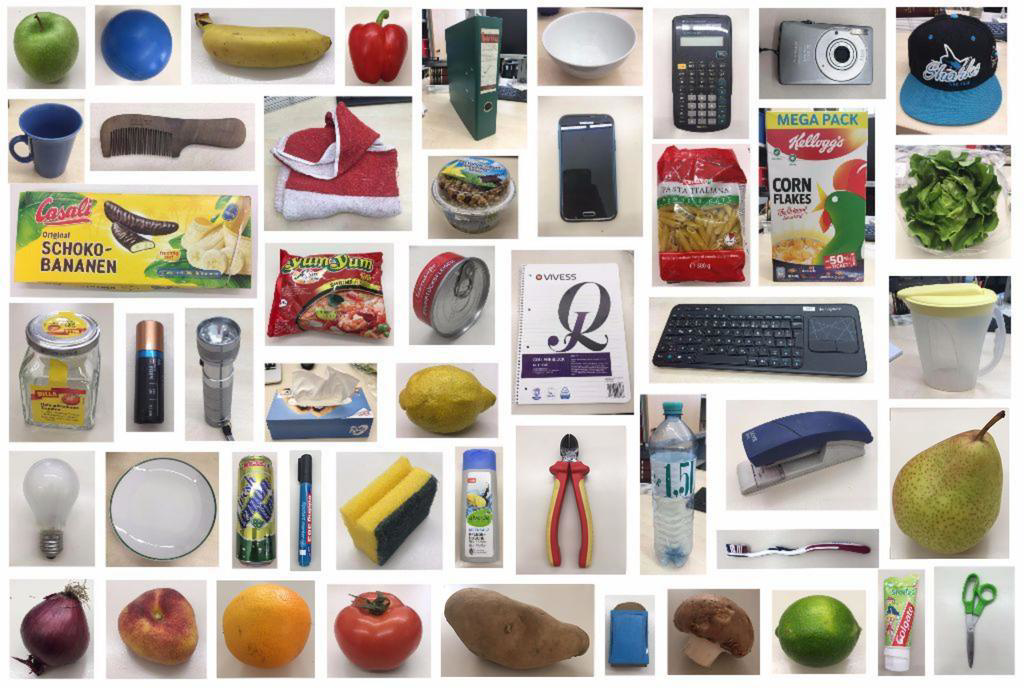}
      \caption{Sample of objects used in Autonomous Robot Indoor Dataset. Each object shown belongs to a different category.}
      \label{fig: objects}
   \end{figure*}

The Autonomous Robot Indoor Dataset contains RGB and depth images of daily life objects belonging to 51 categories. Each object category contains three instances, for a total of 153 physical objects, and it coincides with one of the 51 WordNet leaf nodes used to determine the categories of a very well-known dataset, the RGB-D Object Dataset. In other words, there is a complete overlap between the categories represented in the two datasets, ARID and ROD. Figure~\ref{fig: objects} gives a concrete idea of the dataset's content by showing one sample object per category.

Since we are mostly interested in autonomous assistive robots operating in indoor environments, the object classes considered in ROD are a valid representative. These objects consist of a large variety of food items, such as fruit, vegetables and packed goods, and human-made objects common to homes and offices. Nevertheless, our goal is not to extend and contribute to ROD, but rather fill the gap between research-oriented datasets and real-life data by introducing a robotic dataset collected in-the-wild. While ROD contains images collected in a constrained setting (fixed camera-object distance, static background, invariant light conditions), our dataset includes all the nuisances of robotic data by acquiring it directly with a mobile robot navigating autonomously in an indoor environment. More precisely, the following challenges are taken into account:
\begin{itemize}
\item variation of lighting conditions,
\item object scale variation,
\item significant changes in the viewpoint,
\item partial view and occlusion,
\item clutter, and 
\item background variation.
\end{itemize}
We hope that this work provides the robot vision community with a tool to advance the visual capabilities of robots in order to accelerate their integration in our lives.

\subsection{Data Acquisition Protocol}
In order to avoid a human bias in data acquisition and to observe the objects from the robot's perspective, a mobile robot with an RGB-D camera is used. In particular, the mobile robotic platform is powered by a Pioneer P3-DX with a customized structure that supports an Asus Xtion Pro camera mounted on a pan/tilt unit (see figure~\ref{fig: v4core}).

The data collection is performed in 10 different sessions conducted during different days and at different times of the day: this allows a natural variation of the lighting conditions among the data. At each run, 30-31 objects are spread in the environment where the mobile robot patrols predefined waypoints. When a waypoint is reached, the camera scans the scene with a horizontal movement of the pan/tilt unit and acquires RGB and depth data, both with a resolution of 640x480 pixels and a frame rate of 30 Hz. The RGB and the depth frames are later synchronized based on their acquisition time and unmatched frames are discarded. Each session lasts for approximately one hour in which the robot continuously loops over four distinct waypoints. In order to guarantee the appropriate variability in terms of camera-object distance and object view, the objects are randomly moved in between two patrolling loops.

\subsection{Annotation}
\begin{figure}[t]
      \centering
      \includegraphics[width=0.45\textwidth]{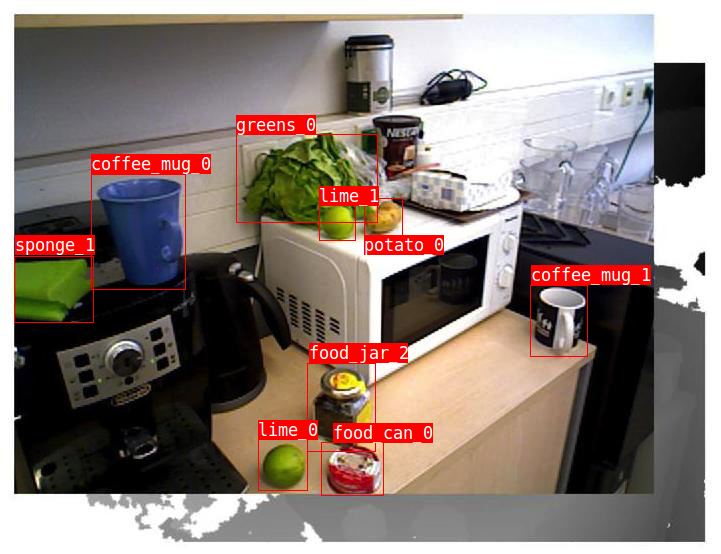}
      \caption{Example of an RGB-D frame from the Autonomous Robot Indoor Dataset with 2D bounding box annotation.}
      \label{fig: frame}
   \end{figure}

In order to discard similar frames, every fifth frame is chosen for annotation for a total of over 6,000 frames. For each frame, a bounding box annotation indicates the location and the label (at instance level) of every visible object for a total of over 120,000 2D bounding boxes for the whole dataset. A modified version of Sloth annotation tool \cite{sloth} is used for this purpose. In case of occlusion or partial view, if the object is still distinguishable, a bounding box is drawn around the visible part of the object. Figure~\ref{fig: frame} shows a sample frame, together with its bounding box annotation. Since the objects are captured in a realistic scenario rather than in isolation, the dataset is also suitable for object detection. In addition, the availability of object labels at instance level allows the dataset to be used for object classification as well as object identification (also referred to as instance recognition).  

\section{Experiments}
\label{sec: experiments}

We take advantage of the availability of ARID to disclose the characteristics of robotic data. In particular, we want to (i) analyze the transferability of features from the Web domain to the robotic domain (Section \ref{subsec: exp_features})  and (ii) study the characteristics of robotic data to identify the main source(s) of complication for classifying objects (Section \ref{subsec: exp_small}). In order to accomplish these goals, another dataset, called Web Object Dataset (WOD), is collected. WOD is composed of images downloaded from the Web representing objects from the same categories as ARID. The images are downloaded from multiple search engines (Google, Yahoo, Bing and Flickr) using the method proposed by Massouh et al.~\cite{web_download}. This method uses a concept expansion strategy by leveraging visual and natural language processing information to minimize the noise while maximizing the visual variability. The remaining noise is then manually removed, leaving a total of 50,547 samples. 

\subsection{Baseline and Features Transferability}
\label{subsec: exp_features}

   \begin{figure}[t]
      \centering
      \includegraphics[width=0.45\textwidth]{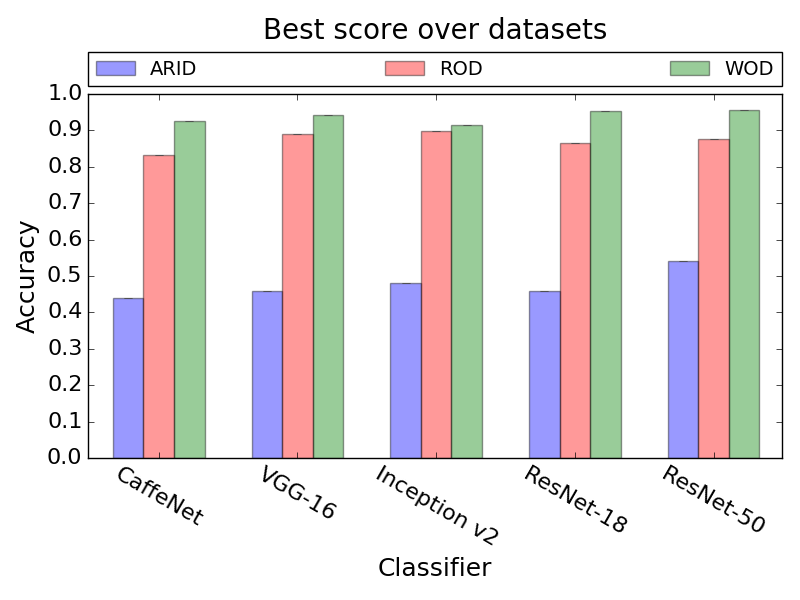}
      \caption{Accuracy of different deep convolutional networks on three datasets: Autonomous Robot Indoor Dataset (ARID), RGB-D Object Dataset (ROD)~\cite{wrgb-d} and Web Object Dataset (WOD). The results are obtained by training and testing on different splits of the same dataset.}
      \label{fig: same_ds}
   \end{figure}

The limited availability of robotic data raises the question of whether data coming from a more accessible domain, the Web domain, can be effectively used instead of data from the RGB-D domain to learn features that are transferable to the robotic data. In particular, we want to compare the performance of well-known deep convolutional networks on robotic data (ARID), when trained on Web data (WOD) and on RGB-D data (ROD). In order to allow a fair evaluation, a subset of 40,000 samples from ARID dataset is selected, such that all the involved datasets are approximately the same size. It is worth noticing that, since WOD does not contain depth information, only RGB data are considered for all datasets. For this benchmark, we employ some of the most utilized network architectures in the literature, CaffeNet\footnote{A slightly modified version of AlexNet in which the normalization is performed after the pooling.}, VGG-16, Inception v2, ResNet-18 and ResNet-50. All networks are pre-trained on ImageNet and then fine-tuned on the desired dataset, according to the guidelines provided in~\cite{ft1}~\cite{ft2}.

   \begin{figure}[t]
      \centering
      \includegraphics[width=0.45\textwidth]{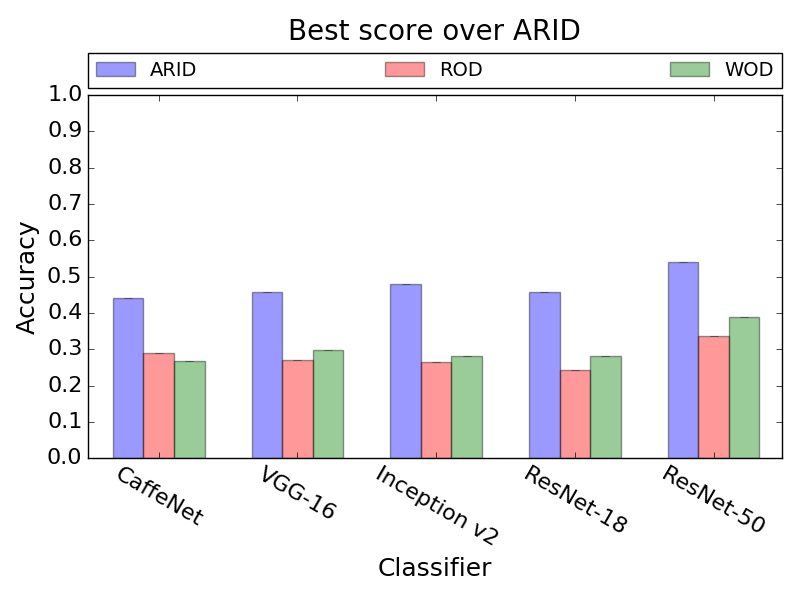}
      \caption{Accuracy of different deep convolutional networks on Autonomous Robot Indoor Dataset (ARID). The results are obtained by training independently on ARID, RGB-D Object Dataset (ROD)~\cite{wrgb-d} and Web Object Dataset (WOD) and testing on ARID.}
      \label{fig: test_arid}
   \end{figure}

In order to provide a reference for the upcoming evaluations, we assess the performances of all considered networks for each of the three datasets (ARID, ROD, WOD) when training and test set come from the same dataset. For each dataset, multiple training/test splits are considered and the results are averaged to obtain the final classification accuracy. In particular, for ARID, each split uses one different object instance per class in the test set, for ROD, the first three splits indicated by the authors is used and, for WOD, each split uses $25\%$ of the data in the test set. From the results displayed in figure~\ref{fig: same_ds}, it can be noticed that the different networks consistently obtain a higher accuracy on WOD. Unsurprisingly, ARID appears to be the most challenging dataset and all the networks achieve an accuracy much lower (on average, $\sim0.4$ lower) on ARID than on the other two datasets. 

\begin{table*}[t!]
\captionof{table}{Accuracy of multiple deep convolutional networks on different training/test combination of three datasets: Autonomous Robot Indoor Dataset (ARID), RGB-D Object Dataset (ROD)~\cite{wrgb-d} and Web Object Dataset (WOD). For each training/test set combination, the mean and maximum accuracy among the considered networks is shown.}
\label{tab: summary}
\centering
\bgroup
\def\arraystretch{1.5}%  1 is the default, change whatever you need
\begin{tabular}{@{}*9l@{}}
 \toprule[1.5pt]
\multicolumn{2}{c}{\head{Dataset}}
& \multicolumn{5}{c}{\head{Network}}
& \multicolumn{2}{c}{\head{Statistics}}\\
 \head{Train on} & \head{Test on}
 & \head{CaffeNet} & \head{VGG-16} & \head{Inception-v2} & \head{ResNet-18} & \head{ResNet-50} 
 & \head{Mean} & \head{Max} \\
 \cmidrule(r){1-2}\cmidrule(l){3-7}\cmidrule(l){8-9}
 ROD & ROD & 
 0.832 & 0.889 & 0.897 & 0.864 & 0.876 &
 \rmfamily 0.872 & 0.897\\
  ROD & ARID & 
 0.291 & 0.270 & 0.266 & 0.243 & 0.337 &
 0.281 & 0.337\\
 WOD & WOD & 
 0.924 & 0.942 & 0.914 & 0.953 & 0.956 &
 0.938 & 0.956\\
  WOD & ARID & 
 0.268 & 0.297 & 0.282 & 0.282 & 0.388 &
 0.303 & 0.388\\
  ARID & ARID & 
 0.441 & 0.458 & 0.481 & 0.458 & 0.540 &
 \rmfamily 0.476 & 0.540\\
 \bottomrule[1.5pt]
\end{tabular}
\egroup
\end{table*}
   
The networks fine-tuned on ROD and WOD are then tested on ARID to evaluate the transferability of the learned features to the robotic data. From the results displayed in figure~\ref{fig: test_arid}, it can be noticed that, as expected, all the networks undergo a performance drop when the training and test set belong to different datasets with respect to the case in which both sets belong to the same dataset (see figure~\ref{fig: same_ds}). The domain shift responsible for this negative inflection of the classification results occurs because the data composing training and test set are drawn from different distributions \cite{domain_adaptation}. However, features learned from Web data (WOD) consistently allow a higher classification accuracy (with improvements up to $0.05$) on robotic data (ARID) than features learned from RGB-D data (ROD) on all networks, with the exception of CaffeNet. The key factor to interpret this phenomenon is the greater variability of Web images: while ROD contains a limited number of instances per class, with some classes containing only 3 instances, in WOD each sample potentially represents a different object instance. Very deep networks, like ResNet-50, with high capacity and generalization power, take advantage of this richness in information to generate better models. This is further highlighted by the difference between the accuracy of ResNet-50 and the mean accuracy of all tested networks when training with WOD (see table~\ref{tab: summary}). The results of this experiment have a twofold implication: (i) despite the greater visual affinity between the RGB-D and the robotic domain, data from the Web domain generate more effective models for object classification in robotic applications, and (ii) the currently well-established deep convolutional network, when used in their plain stand-alone form and without any prior, do not perform satisfactorily for object classification in robotics.

\subsection{Robotic Challenges}
\label{subsec: exp_small}

In order to better understand which characteristics of robotic data negatively influence the results of the object classification task, we independently analyze three key variables: image dimension, occlusion and clutter\footnote{Since ARID is collected in-the-wild, by definition, the data acquisition is performed in an unconstrained manner. For this reason, rigorously isolating other characteristics of the data, such as light variation, background variation and different object view is prohibitive.}. Image dimension is a variable related to the camera-object distance: when the camera is not near enough to clearly capture the object, the object occupies only few pixels in the whole frame, making the classification task more challenging. For obvious reasons, this problem is emphasized when dealing with small and/or elongated objects, such as dry batteries or glue sticks. Occlusion occurs when a portion of an object is hidden by another object or when only part of the object enters the field of view. Since distinctive characteristics of the object might be hidden, occlusion makes the classification task considerably more challenging. Clutter refers to the presence of other objects in the vicinity of the considered object. The simultaneous presence of multiple objects may interfere with the classification task.

\begin{table}[t]
\captionof{table}{Accuracy of ResNet-50, trained on Web Object Dataset and on its augmented version (++), for three subsets of Autonomous Robot Indoor Dataset containing small images, occluded objects and clutters. The model is also tested on the whole dataset to show the overall impact of data augmentation.}
\label{tab: challenges}
\centering
\bgroup
\def\arraystretch{1.5}%  1 is the default, change whatever you need
%\begin{tabular}{@{}*3l@{}}
\begin{tabular}{m{2.5cm} m{1.5cm} m{1.5cm}}
 \toprule[1.5pt]
\multicolumn{1}{c}{\head{Challenge}}
& \multicolumn{2}{c}{\head{Accuracy}}\\
 \head{}
 & \head{Top-1} & \head{Top-5}\\
 \cmidrule(r){1-1}\cmidrule(l){2-3}
 Small image &
 0.230 & 0.511\\
 Occlusion &
 0.273 & 0.508\\
 Clutter &
 0.558 & 0.777\\
   Small image ++ &
 0.240 & 0.513\\
 Occlusion ++ &
 0.318 & 0.577\\
   Clutter ++ &
 0.543 & 0.802\\\hline
  All ++ &
 0.441 & 0.702\\
 \bottomrule[1.5pt]
\end{tabular}
\egroup
\end{table}

   \begin{figure*}[th]
      \centering
      \includegraphics[width=0.9\textwidth]{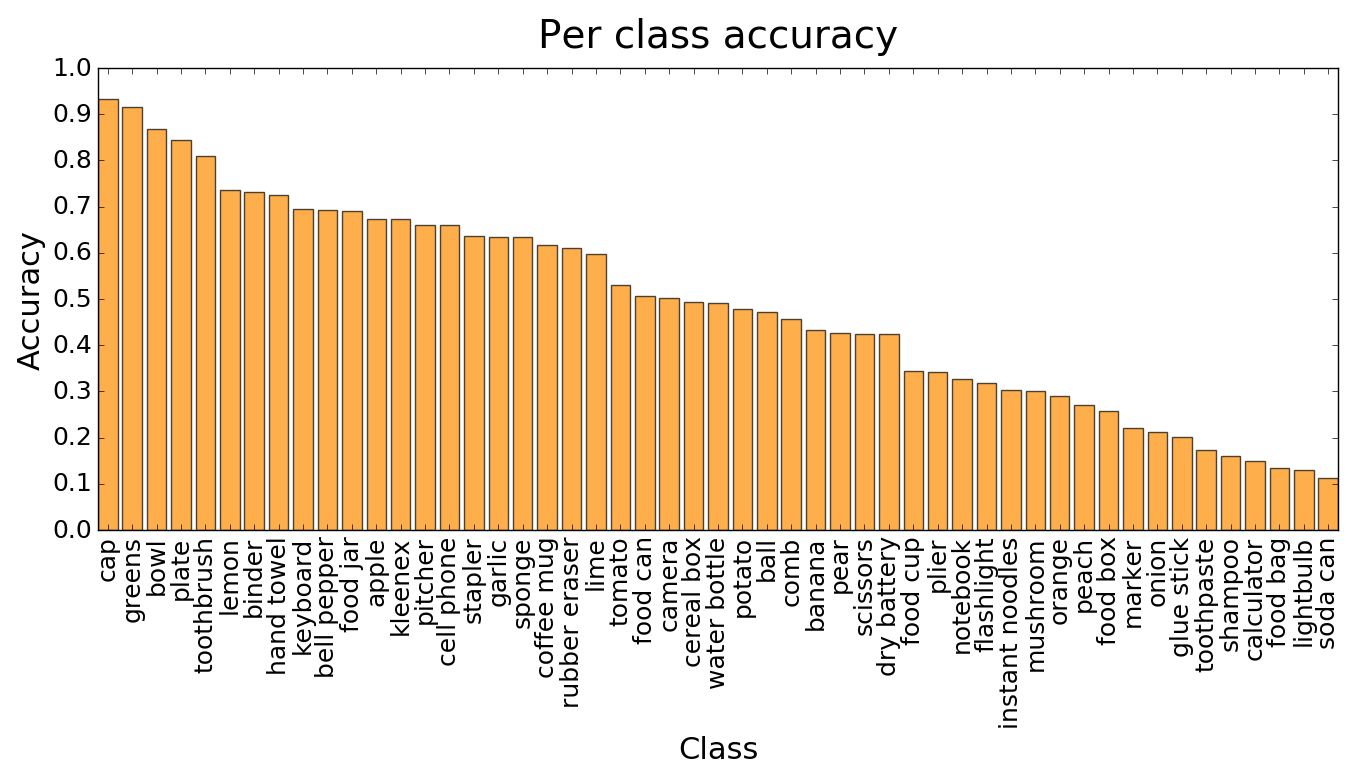}
      \caption{Accuracy of each of the 51 classes of the Autonomous Robot Indoor Dataset obtained with a ResNet-50 trained on the augmented Web Object Dataset.}
      \label{fig: per_class}
   \end{figure*}

Table~\ref{tab: challenges} shows the classification results of the best performing model of section \ref{subsec: exp_features} (ResNet-50 trained on WOD) on three subsets of ARID, each containing samples with the characteristics discussed above. The set of small images is obtained by taking half of ARID containing images with the smallest area, while the occlusion and clutter set have been manually selected. It is worth noticing that the three sets are mutually exclusive in order to avoid interference between the analyzed variables. The occlusion and, especially, the small images set exhibit low accuracy, thus negatively affecting the classification score of the whole dataset. It is possible to improve the classification by performing problem-specific data augmentation during the training phase. In particular, we augmented WOD by resizing the original samples to different scales and by randomly adding rectangular noise patches to simulate occlusion. These two strategies are commonly used to encourage the network to learn scale-/occlusion-invariant models \cite{inception} \cite{depthnet} \cite{dilated}. Table~\ref{tab: challenges} also shows the performances of ResNet-50 trained with this augmented WOD on the three subsets and on the whole ARID dataset. Even though the occlusion set benefits from this strategy (and so does the whole dataset) the classification of small images does not exhibit significant improvement. The difficulty of classifying small images is further confirmed by the results in figure~\ref{fig: per_class}, where classes representing small or elongated objects have the lowest accuracy.

\section{Discussion and Conclusion}
\label{sec: conclusions}

In this paper, we have presented ARID: a large-scale, multi-view, RGB-D object dataset collected with a mobile robot in-the-wild. This dataset is designed to capture the challenges a robot faces when deployed in an indoor environment and fills the current gap in the robot vision community between research oriented datasets and real-life data. Furthermore, with an extensive comparative study, we have shown that it is possible to overcome the complication of collecting a large amount of robotic data for training data-craving deep convolutional networks by using images downloaded from the Web. We have found that, despite being relatively easy to obtain, Web-based data allow the generation of more effective deep models than the RGB-D counterpart for the classification of robotic images. Nevertheless, object classification remains a challenging task in robotics and current algorithms present results that are insufficient for a successful integration of robotic systems in our homes. In order to shed light on the difficulties of this task, we have analyzed the effects of specific factors, such as object dimension, occlusion and clutter, on the performance. Results indicate that clutter is rather a secondary problem: occlusions and, especially, small objects more seriously degrade the classification accuracy. These observations suggest a research path in which visual tasks for robotic applications are tackled through methods designed to cope with domain-specific challenges. ARID is a valuable resource to pursue this goal and provides an important testbed for the robot vision community. In addition, the dataset may also be used to explore other aspects of robotic data, such as the integration of RGB and depth information.

Our dataset is available for download at \url{https://www.acin.tuwien.ac.at/en/vision-for-robotics/software-tools/autonomous-robot-indoor-dataset/}.

\balance

%%%%%%%%%%%%%%%%%%%%%%%%%%%%%%%%%%%%%%%%%%%%%%%%%%%%%%%%%%%%%%%%%%%%%%%%%%%%%%%%

%%%%%%%%%%%%%%%%%%%%%%%%%%%%%%%%%%%%%%%%%%%%%%%%%%%%%%%%%%%%%%%%%%%%%%%%%%%%%%%%

%%%%%%%%%%%%%%%%%%%%%%%%%%%%%%%%%%%%%%%%%%%%%%%%%%%%%%%%%%%%%%%%%%%%%%%%%%%%%%%%
%\section*{APPENDIX}

\section*{ACKNOWLEDGMENT}

This work has received funding from the European Union’s Horizon 2020 research and innovation program under the Marie Skłodowska-Curie grant agreement No. 676157, project ACROSSING, by the ERC grant 637076 - RoboExNovo (B.C.), and the CHIST-ERA project ALOOF (B.C.). The authors would like to thank Victor Le\'{o}n, Mirco Planamente and Silvia Bucci for their help during the data collection and annotation process. The authors are also grateful to Tim Patten for the support in finalizing the work and Georg Halmetschlager for the help in the camera calibration process.

%%%%%%%%%%%%%%%%%%%%%%%%%%%%%%%%%%%%%%%%%%%%%%%%%%%%%%%%%%%%%%%%%%%%%%%%%%%%%%%%

\bibliographystyle{IEEEtran}
\bibliography{references}

\end{document}